\documentclass[10pt,letterpaper,twocolumn]{article}
\usepackage[margin=0.75in,columnsep=0.25in]{geometry}
\usepackage[T1]{fontenc}
\usepackage{newtxtext,newtxmath}
\usepackage[hyphens]{url}
\usepackage{graphicx}
\usepackage[authoryear]{natbib}
\usepackage{caption}
\usepackage[hidelinks]{hyperref}
\usepackage{amsmath}
\usepackage{array}
\usepackage{booktabs}
\usepackage{multirow}
\usepackage{textcomp}

\title{Geometric Risk Control for Vision-Language Model OCR}
\author{\textbf{Weile Gong\textsuperscript{1}, Zijian Lu\textsuperscript{2},
Mingcai Chen\textsuperscript{3}, 
 Yiping Zuo\textsuperscript{4},Xin He\textsuperscript{5}, Weibei Fan\textsuperscript{6}}\\[0.3em]
\normalsize Nanjing University of Posts and Telecommunications, Nanjing, China\\
\normalsize \textsuperscript{1}b25011527@njupt.edu.cn;
\textsuperscript{2}18818732360@163.com;
\textsuperscript{3}chenmc@njupt.edu.cn;\\
\normalsize \textsuperscript{4}zuoyiping@njupt.edu.cn;
\textsuperscript{5}xhe@njupt.edu.cn;
\textsuperscript{6}wbfan@njupt.edu.cn
}
\date{}

\begin{document}
\maketitle

\begin{abstract}
Vision-language models (VLMs) enable flexible generative optical character
recognition (OCR), while their open-ended decoders can expose wrong but fluent text with
weak visual support. In audit-sensitive records, such an output can be more
costly than abstention. Frozen or externally served VLMs therefore require an
external decision layer that can determine whether a transcription has
sufficient visual evidence for release. We introduce the Geometric Risk
Controller (GRC), a model-agnostic controller that treats controlled geometric
transformations as repeatable black-box probes, screens structurally
implausible continuations, and releases the unique candidate supported by
coherent cross-view evidence. The protocol provides empirical selective
exposure control with explicit coverage and query cost under a reproducible
fixed decision rule.
Experiments across frozen VLMs and standard scene-text benchmarks consistently
reduce mean, upper-tail, and catastrophic error among released outputs while
retaining high coverage.
\end{abstract}

\noindent\textbf{Code:} \url{https://github.com/phare111/GRC}

\section{Introduction}

Generative OCR places fluent visual text directly into downstream records.
This capability supports financial processing, identity verification, medical
documentation, and safety-critical archives. Frozen VLMs extend the reach of
OCR through open-ended decoding \cite{huang2024kosmos25,yu2024text}, and the
same decoding process can produce over-generation, unsupported substitutions,
and severe continuations when visual evidence is weak. Audit-sensitive systems
therefore require every released string to satisfy an evidence standard that
supports escalation whenever that standard is unmet.

Reliable release requires an instance-level exposure decision. Autoregressive
likelihood measures compatibility with the generated prefix and remains
entangled with the model language prior. Single-view confidence can remain high
on hallucinated outputs \cite{li2023evaluating,mao2025calibrating}, and prompt
or decoding choices can reshape the severe-error tail
\cite{zhao2021calibrate,rudman2026pih,huang2024opera}. Frozen and externally
served backbones also limit access to retraining and internal calibration.
These conditions create a scientific need for evidence obtained from the
visual input through a reproducible black-box protocol.

\begin{figure*}[t]
  \centering
  \includegraphics[width=0.9\textwidth]{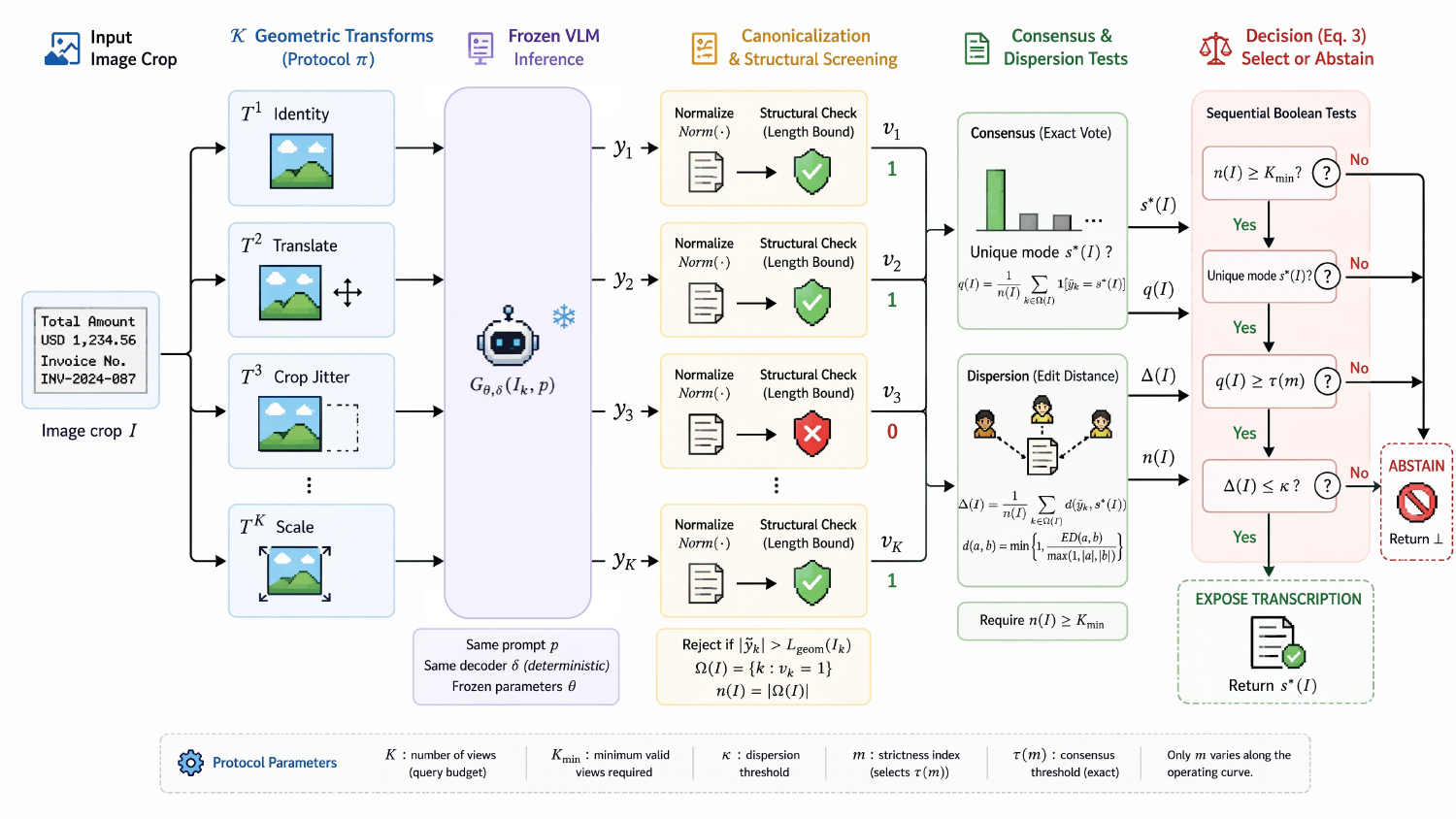}
  \caption{Overview of GRC. Protocol \(\pi\) applies \(K\) transforms
  \(I_k=T^k(I)\). The frozen model \(G_{\theta,\delta}\) with parameters
  \(\theta\), common prompt \(p\), and fixed decoder \(\delta\) returns
  \(y_k\). Canonicalization yields \(\tilde y_k\). Validity flag \(v_k\)
  rejects strings beyond the geometric length bound \(L_{\mathrm{geom}}\).
  The valid set is \(\Omega(I)=\{k\mid v_k=1\}\) with
  \(n(I)=|\Omega(I)|\).
  Valid strings must have unique mode \(s^*(I)\). Exact support \(q(I)\) and
  mean normalized edit-distance dispersion \(\Delta(I)\) then enter
  Eq.~\ref{eq:controller}. The controller exposes \(s^*(I)\) iff
  \(n(I)\ge K_{\min}\), \(q(I)\ge\tau(m)\), and
  \(\Delta(I)\le\kappa\); otherwise return \(\bot\).
  Here \(K_{\min}\) is the minimum valid-view count, \(\tau(m)\) the consensus
  threshold chosen by strictness \(m\), and \(\kappa\) the dispersion cap.
  Only \(m\) varies along the operating curve.}
  \label{fig:teaser}
\end{figure*}

Selective exposure changes the statistical object of interest from an average
recognition score to the error distribution among released outputs. Coverage
records the probability of release, and conditional risk records the severity
of errors that remain visible. Their joint measurement reveals whether a
controller concentrates difficult cases in the escalation stream. This
decomposition is essential for high-consequence OCR because a small severe
tail can dominate downstream cost even when average recognition remains
competitive.

Controlled geometry supplies this evidence through intervention. Small
content-preserving transformations alter pixel alignment while the backbone,
prompt, and deterministic decoder remain fixed. A visually supported
transcription should retain coherent support across these interventions.
Over-generation and brittle substitutions often lose agreement or move far
apart in edit space. This isolation of visual variation turns cross-view
behavior into an external measurement of support and makes every decision
replayable from the observed outputs.

GRC converts the measurement into a selective exposure rule. Figure
\ref{fig:teaser} shows a fixed family of geometric views followed by
canonicalization, structural screening, exact consensus, edit-distance
dispersion, and a sequential decision. The controller releases a unique
candidate after every evidence gate passes and sends the remaining crops to an
escalation path. Protocol parameters define the evidence collection process
once. A strictness index selects a measured coverage and risk operating point
while the backbone, transforms, screening rule, and dispersion threshold stay
fixed.

This design targets the deployment regime where unsupported text carries high
downstream cost and abstention has a useful destination. GRC can serve as a
second-stage controller for uncertain, high-value, or audit-critical crops.
Independent views admit batching and parallel execution, and the query budget
sets an explicit latency envelope. The resulting protocol provides a practical
release decision while leaving the recognition backbone unchanged.

Our experiments evaluate frozen generators under one shared protocol and
measure mean error, upper-tail error, catastrophic exposure, coverage,
component contributions, matched-confidence selection, and query cost. The
results show consistent risk reduction among released outputs. Accepted and
abstained subsets also exhibit a large accuracy separation, which confirms
that GRC concentrates difficult crops in the escalation stream. This evidence
supports three contributions.

\begin{itemize}
  \item We formulate selective exposure as a first-class interface for frozen
  generative OCR in audit-sensitive deployment.
  \item GRC converts protocol-fixed geometric interventions into reproducible
  black-box evidence through complementary structural and consistency tests.
  \item A unified evaluation connects exposed tail risk, coverage, component
  behavior, confidence selection, and query cost under shared operating points.
\end{itemize}

\section{Related Work}

\paragraph{Generative OCR and deployment exposure.}
VLMs such as KOSMOS-2.5, TextHawk, and OCEAN-OCR demonstrate strong
transcription ability across visual-text settings
\cite{huang2024kosmos25,yu2024text,chen2025oceanocrgeneralocrapplication}.
Benchmarks including ICDAR, SVT, and OCRBench emphasize aggregate recognition
\cite{karatzas2013icdar,SVT,ocrbenchv1,fu2025ocrbenchv2}. Deployment reliability
also depends on the severe errors that reach downstream users
\cite{AmitATF,huang2024opera}. Our study centers this released-output
distribution and measures it under a fixed selective protocol.

\paragraph{Selective prediction and formal risk control.}
Selective prediction exposes a coverage and risk trade-off through an
accept/abstain interface \cite{chow1970,geifman2017selectivenet,wen-etal-2025-know}.
Conformal risk control provides finite-sample control for a specified
loss under its assumptions \cite{Angelopoulos2021AGI,angelopoulos2024conformal,
Wang2025SAFERRS}. GRC adopts the selective interface and reports empirical,
protocol-fixed exposure controls. Its evidence comes from controlled visual
interventions because internal confidence and self-evaluation can be
prompt-sensitive or overconfident on hallucinations
\cite{Schmalfuss2025PARCAQ,zollicoffer2025mtremultitokenreliabilityestimation,
woo-etal-2025-dont}.

\paragraph{Test-time multi-view evidence.}
Test-time augmentation and self-consistency commonly aggregate repeated
predictions to improve accuracy \cite{RA-TTA,wang2023selfconsistency,
snell2024scalingllmtesttimecompute,kaya2025efficienttesttimescalingsmall}.
GRC uses repeated inference to measure release evidence. Controlled geometry
isolates image sensitivity, structural constraints suppress malformed strings
\cite{willard2023efficientguidedgenerationlarge}, and cross-view statistics
measure agreement. GRC turns these responses into an external selective release
signal that complements the underlying recognizer.

\section{Geometric Selective Control}

\subsection{Black-Box Exposure Contract}

Let a frozen VLM with fixed prompt \(p\), decoding configuration \(\delta\), and
parameters \(\theta\) map an image crop to a string
\[
G_{\theta,\delta}(I,p)\in\mathcal{Y}.
\]
GRC preserves this generator and adds a selective map with outputs in
\(\mathcal{Y}\cup\{\bot\}\), where \(\bot\) denotes abstention. Protocol
\(\pi\) produces a candidate together with external evidence. The controller
then maps each crop to a released transcription or an escalation decision.
This contract makes exposure a measurable system output and leaves the
recognition backbone unchanged.

\subsection{External Evidence from Controlled Geometry}

Controlled geometry creates an isolated visual intervention. The generator,
prompt, and decoder remain fixed while the input pixel arrangement changes
through mild content-preserving transformations. Agreement across these views
therefore reflects response stability under visual intervention. This
measurement remains available through a black-box inference interface and is
independent of architecture-specific confidence calibration.

Protocol \(\pi\) constructs a finite intervention set
\[
\mathcal{T}_{\pi}(I)=\{T^k(I)\}_{k=1}^{K}.
\]
Content preservation keeps the intended transcription invariant across this
set. The observed string sequence then forms an intervention response profile.
Exact agreement measures concentrated support, edit dispersion measures the
severity of residual variation, and the valid-view count measures how much
structural evidence survives. These quantities separate the decision from a
single decoder score and give every release a concrete evidential trace.

The intervention targets failures with weak geometric support.
Over-generation, content-free continuation, and visually brittle
substitutions often change when translation, crop jitter, or scale perturbs
the evidence received by the visual encoder. Their instability supplies a
direct abstention signal under the declared operating point.

Abstention preserves the evidence boundary after a failed gate. Every
disagreeing alternative lacks the support required for release under the same
protocol. The surrounding system can route the crop to human review, a
stronger recognizer, or an independent verifier. This separation keeps
recognition and exposure decisions explicit and reproducible.

\subsection{Protocol-Fixed Geometric Evidence}

Protocol \(\pi\) contains \(K\) content-preserving transforms. They include the identity
\(T^1\) and mild translations, crop jitters, and scale variations
\(T^2,\ldots,T^K\). Every view uses the same prompt and deterministic decoder
\[
I_k=T^k(I),\qquad y_k=G_{\theta,\delta}(I_k,p),\quad k=1,\ldots,K.
\]
The transforms and all controller thresholds are fixed before evaluation and
shared across backbones and datasets. Thus improvements cannot arise from
per-model or per-dataset protocol tuning.

The evidence reduction follows two complementary stages. Canonicalization
removes presentation variation that carries no transcriptional meaning.
Structural screening then identifies continuations whose form conflicts with
the geometric support of the crop. This ordering protects the consensus test
from superficial string variation and from runaway outputs that could
otherwise dominate edit statistics. The remaining strings constitute a
protocol-defined evidence set shared by support and dispersion tests.

Each raw output is mapped to
\(\tilde y_k=\mathrm{Norm}(y_k)\), where the fixed normalization removes only
presentation variation such as repeated whitespace and task-appropriate case.
A label-free structural indicator \(v_k\) screens degenerate continuations.
Its principal check is a permissive geometric length bound
\(L_{\mathrm{geom}}(I_k)\) derived from normalized foreground geometry and
dominant-axis span. The rule sets \(v_k=0\) when
\(|\tilde y_k|>L_{\mathrm{geom}}(I_k)\). The bound serves as an
over-generation guard and keeps font or crop sensitivity visible through
coverage. Valid evidence is indexed by
\(\Omega(I)=\{k\mid v_k=1\}\), with \(n(I)=|\Omega(I)|\).

\paragraph{Separation of protocol and operating point.}
The protocol parameters have distinct and testable roles. \(K\) specifies the
query budget. \(K_{\min}\) specifies the minimum evidence count. \(\kappa\)
caps dispersion around the candidate. \(m\) selects the exact-consensus level
\(\tau(m)\). Transform sampling, normalization, structural screening,
\(K_{\min}\), and \(\kappa\) remain fixed along the reported operating curve.
Only \(m\) changes the release strictness. This separation preserves one
evidence definition across models and datasets and exposes every rejected crop
through the measured coverage.

The separation also induces nested release sets. Define
\(\mathcal{A}_{\pi,m}=\{I\mid f_{\pi,m}(I)\neq\bot\}\). For two strictness
levels with \(\tau(m_1)\le\tau(m_2)\), the fixed remaining gates give
\[
\mathcal{A}_{\pi,m_2}\subseteq\mathcal{A}_{\pi,m_1}.
\]
Increasing the consensus threshold can remove releases and cannot introduce a
new release. This deterministic ordering gives \(m\) one interpretable role
and turns the reported operating curve into a family generated from a single
evidence protocol.

\subsection{Consensus, Dispersion, and Decision}

When \(n(I)\ge K_{\min}\), let \(s^*(I)\) be the unique mode among valid
canonical strings. Absence of a unique mode produces abstention. Exact support
and residual disagreement are
\begin{align}
q(I) &=
\frac{1}{n(I)}\sum_{k\in\Omega(I)}
\mathbf{1}\!\left[\tilde y_k=s^*(I)\right],\\
\Delta(I) &=
\frac{1}{n(I)}\sum_{k\in\Omega(I)}
d\!\left(\tilde y_k,s^*(I)\right),
\end{align}
where
\(d(a,b)=\min\{1,\mathrm{ED}(a,b)/\max(1,|a|,|b|)\}\).
Normalization handles presentation-only variants. Edit dispersion records
near-matches while distinct transcriptions remain separate observations.

The statistics capture different properties of the evidence distribution.
\(q(I)\) measures empirical mass at the modal string, while \(\Delta(I)\)
measures average residual displacement from that string. Two crops can share
the same modal support and exhibit very different residual errors. Joint
testing therefore distinguishes a compact cluster from a majority surrounded
by distant alternatives. The unique-mode gate adds identifiability and removes
release decisions produced by tied evidence.

For strictness index \(m\), consensus threshold \(\tau(m)\), and fixed
dispersion threshold \(\kappa\), the controller is
\begin{equation}
f_{\pi,m}(I)=
\begin{cases}
s^*(I),&
\begin{array}{l}
n(I)\ge K_{\min},\ s^*(I)\text{ unique},\\[-1pt]
q(I)\ge\tau(m),\ \Delta(I)\le\kappa,
\end{array}\\
\bot,&\text{otherwise.}
\end{cases}
\label{eq:controller}
\end{equation}
Increasing \(m\) changes only the required consensus and yields a more
conservative operating point. Equation~\ref{eq:controller} controls evidence
stability under the declared protocol.

\begin{figure}[t]
  \centering
  \includegraphics[width=\columnwidth]{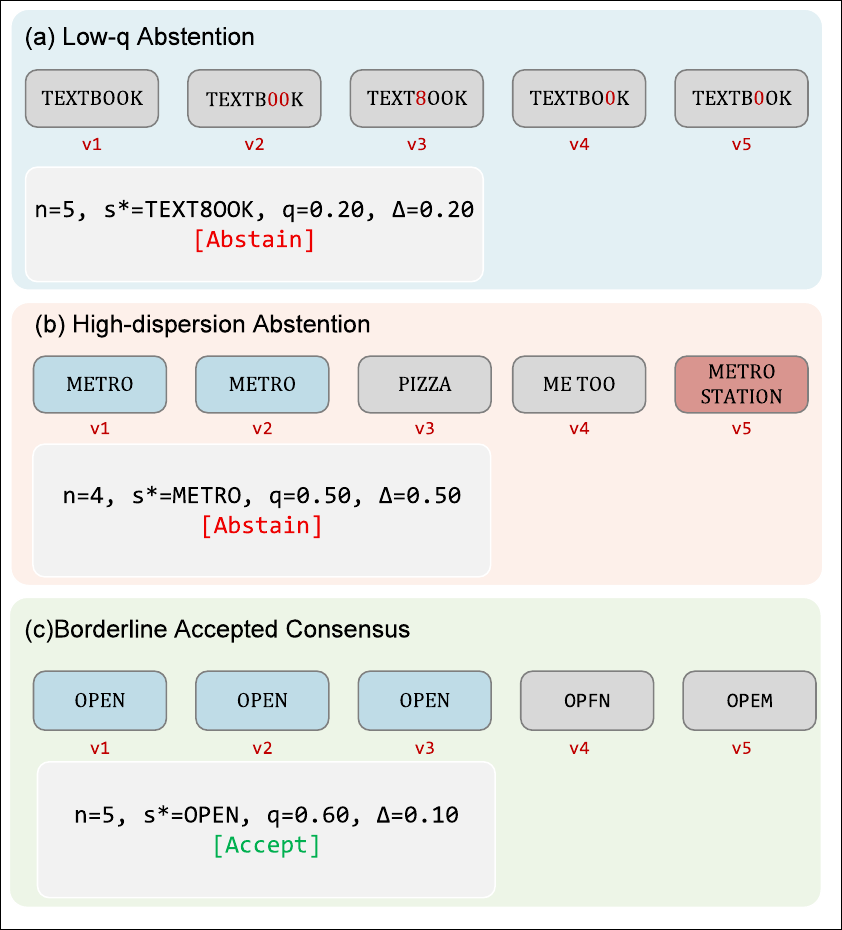}
  \caption{Evidence patterns for \(K=5,m=3\). Fragmented support abstains. A
  majority may still abstain when dispersion is high. Tightly clustered
  evidence can pass at moderate support.}
  \label{fig:evidence_patterns}
\end{figure}

The unique-mode requirement removes arbitrary tie breaking from release
decisions. Exact voting follows normalization, so presentation variation is
absorbed and visually meaningful character changes remain disagreements. The
dispersion gate complements this discrete statistic. A nominal majority with
distant residual strings produces large \(\Delta\) and fails the gate. A
mode with moderate support and tightly clustered residual strings can pass. Figure
\ref{fig:evidence_patterns} illustrates these evidence geometries.

\subsection{Coverage and Exposed Risk}

Coverage is
\(\mathrm{Cov}(\pi,m)=\Pr[f_{\pi,m}(I)\neq\bot]\).
All controlled-system risk is conditional on this covered subset. Given ground
truth \(y^{\mathrm{gt}}\), we report mean CER, 99th-percentile CER (P99), and
\[
\mathrm{MD}@2=
\Pr[\mathrm{CER}(f_{\pi,m}(I),y^{\mathrm{gt}})\ge2
\mid f_{\pi,m}(I)\neq\bot].
\]
Because CER is normalized by ground-truth length, runaway generation can exceed
one. MD@2 uses per mille units (\textperthousand).

This conditional definition is important. Baseline risk is measured on all
samples because the baseline always releases a result. GRC risk is measured on
released outputs, and coverage gives the probability of release. The system
objective concentrates difficult cases in the escalation set and reduces the
severity of visible errors. Reporting coverage beside every conditional risk
keeps this concentration measurable.

\section{Experiments}

The evaluation tests four linked claims under one shared protocol. Geometric
intervention should reduce the mean and severe tail of error among released
outputs. External cross-view evidence should remain informative at coverage
matched to an internal confidence selector. Structural and consensus evidence
should contribute complementary reductions in catastrophic exposure. The
query-budget sweep should identify a practical saturation point for the
second-stage deployment role. Main results, matched selection, component
analysis, and cost analysis test these claims in this order.

\subsection{Setup}

\paragraph{Backbones and data.}
We use frozen LLaVA-Phi3 (3.8B) \cite{llavaphi3}, Gemma3 (4B)
\cite{gemma3}, and GLM-OCR \cite{glmocr} as black-box generators on the
official IIIT5K \cite{IIIT5K} and ICDAR 2013 \cite{karatzas2013icdar} test
sets. Their word and short-crop labels provide exact ground truth for
controlled coverage and exposed-risk accounting.

\paragraph{Fixed protocol and operating points.}
The default uses one anchor and four transformed views (\(K=5\)),
\(K_{\min}=3\), and \(\kappa=0.4\). Predefined strictness levels
\(m\in\{1,3,5\}\) use
\(\tau(1)=0.1,\tau(3)=0.5,\tau(5)=0.9\), shared across all six settings.
Unless noted, results use \(m=3\). Baseline outputs receive the same
canonicalization as GRC. Every sample remains in the evaluation, and one
controller specification is shared by every backbone and dataset.

\begin{figure*}[t]
  \centering
  \includegraphics[width=\textwidth]{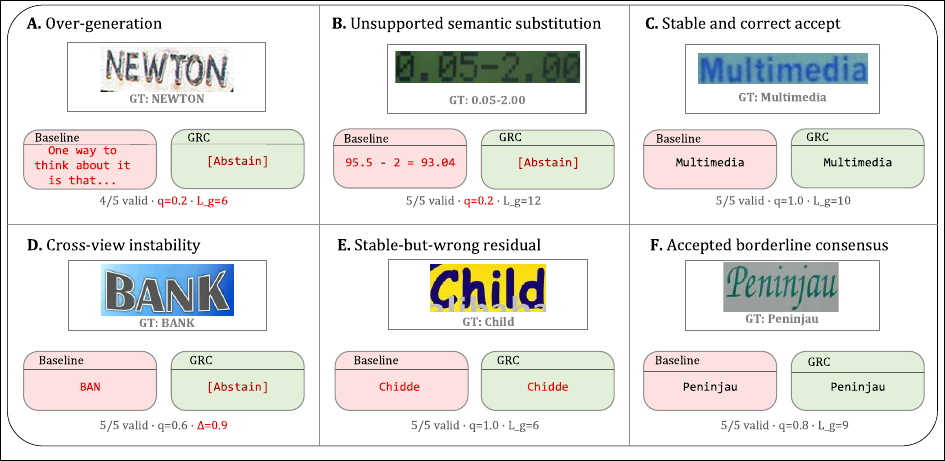}
  \caption{Representative fixed-protocol outcomes at \(m=3\). GRC abstains on
  over-generation, unsupported substitution, and view-sensitive guesses,
  while retaining supported cases across different consensus levels. Evidence
  tags report valid views, consensus \(q\), and \(L_{\mathrm{geom}}\), or
  dispersion \(\Delta\) for the instability case.}
  \label{fig:qual_cases}
\end{figure*}

\subsection{Exposed-Risk Reduction}

\begin{table*}[t]
  \centering
  \small
  \setlength{\tabcolsep}{3.5pt}
  \begin{tabular}{llrrrrrrrr}
    \toprule
    & & \multicolumn{4}{c}{IIIT5K} & \multicolumn{4}{c}{ICDAR13}\\
    \cmidrule(lr){3-6}\cmidrule(lr){7-10}
    Model & Method & Cov. & CER & P99 & MD@2 & Cov. & CER & P99 & MD@2\\
    \midrule
    \multirow{2}{*}{LLaVA-Phi3}
      & Baseline & 100.0 & 110.5 & 3087.8 & 33.7 & 100.0 & 147.4 & 3572.5 & 35.6\\
      & GRC      & 89.5  & \textbf{8.4} & \textbf{100.0} & \textbf{0.3}
                 & 89.0 & \textbf{9.3} & \textbf{100.0} & \textbf{3.7}\\
    \multirow{2}{*}{Gemma3}
      & Baseline & 100.0 & 7.4 & 100.0 & 3.7 & 100.0 & 20.1 & 500.0 & 16.4\\
      & GRC      & 94.9 & \textbf{3.7} & \textbf{66.7} & \textbf{0.0}
                 & 93.0 & \textbf{3.7} & \textbf{100.0} & \textbf{1.8}\\
    \multirow{2}{*}{GLM-OCR}
      & Baseline & 100.0 & 6.3 & 100.0 & 3.7 & 100.0 & 3.3 & 88.3 & 1.8\\
      & GRC      & 95.7 & \textbf{4.0} & \textbf{66.7} & \textbf{0.7}
                 & 96.2 & \textbf{2.3} & \textbf{58.0} & \textbf{0.9}\\
    \bottomrule
  \end{tabular}
  \caption{Main results for \(K=5,m=3\). Coverage and CER are percentages.
  MD@2 is per mille. Baseline risk uses the full set. GRC risk uses only exposed
  outputs.}
  \label{tab:main_results}
\end{table*}

Table~\ref{tab:main_results} shows the same pattern in all six
model and dataset settings. GRC retains 89.0 to 96.2\% coverage while reducing mean
CER, P99, and MD@2 relative to always accepting. The LLaVA-Phi3 units are
especially important. A value of 33.7 MD@2 means
33.7\textperthousand\ (3.37\%), and
0.3 means 0.3\textperthousand\ (0.03\%). On 3,000 IIIT5K crops, this is about
101 catastrophic baseline outputs versus about one catastrophic exposed GRC
output, so concentrating failures in the 10.5\% abstained subset is
arithmetically consistent. Figure~\ref{fig:qual_cases} shows representative
release and abstention outcomes.

The directional consistency across backbones is central to the model-agnostic
claim. The evaluated generators begin with very different mean and tail error
profiles, yet the same protocol lowers every reported released-output risk
measure. The controller therefore acts on a shared behavioral signal created
by geometric intervention and avoids dependence on a model-specific confidence
scale.

\paragraph{Abstention set analysis.}
The rejected set contains markedly harder crops. For LLaVA-Phi3 on IIIT5K,
exact-match accuracy is 76.69\% on
accepted samples and 13.22\% on abstained samples, a 63.47-point gap. Together
with 89.5\% coverage, this split recovers the approximately 70.0\% full-set
accuracy of the underlying frozen recognizer. GRC partitions the same test
population into a substantially more reliable released subset and a much
harder escalation subset. The measured coverage makes the corresponding
withholding cost explicit.

\subsection{Comparison at Similar Coverage}

We compare with a confidence-threshold selective baseline adapted from the
vanilla selective component of \cite{srinivasan2024selectiveselectivepredictionreducing}.
It uses the same frozen backbone, prompt, deterministic decoding, and
canonicalization with one view. A held-out split calibrates mean token
log-probability to target GRC's \(m=3\) coverage. Test coverage is therefore
approximately matched.

\begin{table*}[t]
  \centering
  \small
  \setlength{\tabcolsep}{5pt}
  \begin{tabular}{llrrrrrr}
    \toprule
    & & \multicolumn{3}{c}{IIIT5K} & \multicolumn{3}{c}{ICDAR13}\\
    \cmidrule(lr){3-5}\cmidrule(lr){6-8}
    Model & Method & Cov. & CER & MD@2 & Cov. & CER & MD@2\\
    \midrule
    \multirow{2}{*}{LLaVA-Phi3}
      & Conf.-Thr. & 90.0 & 30.9 & 19.6 & 88.2 & 82.7 & 31.4\\
      & GRC        & 89.5 & \textbf{8.4} & \textbf{0.3}
                   & 89.0 & \textbf{9.3} & \textbf{3.7}\\
    \multirow{2}{*}{Gemma3}
      & Conf.-Thr. & 95.8 & 7.6 & 2.2 & 95.5 & 22.5 & 15.1\\
      & GRC        & 94.9 & \textbf{3.7} & \textbf{0.0}
                   & 93.0 & \textbf{3.7} & \textbf{1.8}\\
    \multirow{2}{*}{GLM-OCR}
      & Conf.-Thr. & 94.8 & 4.4 & 2.2 & 95.9 & 3.2 & 5.8\\
      & GRC        & 95.7 & \textbf{4.0} & \textbf{0.7}
                   & 96.2 & \textbf{2.3} & \textbf{0.9}\\
    \bottomrule
  \end{tabular}
  \caption{Approximately coverage-matched confidence comparison. Coverage and
  CER are percentages. MD@2 is per mille.}
  \label{tab:confidence}
\end{table*}

At similar coverage, GRC yields lower mean CER and MD@2 in all six settings
(Table~\ref{tab:confidence}). This supports the use of external perturbation
evidence for selective exposure. The shared coverage range isolates evidence
quality from the amount of released output.

The separation is strongest where the frozen generator has the heaviest tail.
On IIIT5K with LLaVA-Phi3, confidence selection retains 90.0\% coverage with
19.6\textperthousand\ MD@2, while GRC retains 89.5\% coverage with
0.3\textperthousand. On ICDAR13, the corresponding values are
31.4\textperthousand\ and 3.7\textperthousand\ at closely matched coverage.
Gemma3 and GLM-OCR show the same direction on both datasets. The consistency
across baseline error scales supports controlled geometry as a transferable
release signal.

\subsection{Operating Points and Components}

\begin{figure*}[t]
  \centering
  \includegraphics[width=\textwidth]{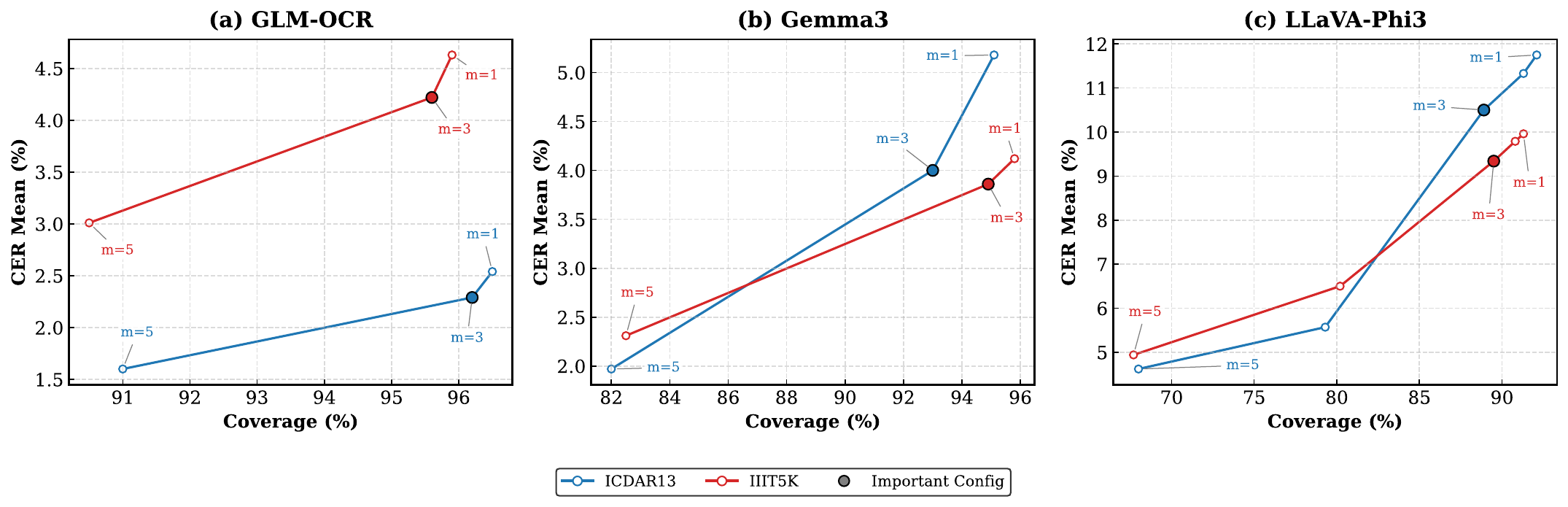}
  \caption{Coverage and risk trajectories under the shared protocol. Each point
  is a predeclared strictness \(m\). Filled markers denote \(m=3\). Increasing
  strictness consistently trades coverage for lower covered-output CER.}
  \label{fig:risk_coverage}
\end{figure*}

Figure~\ref{fig:risk_coverage} shows that \(m\) selects an operating point
without modifying the backbone, transforms, or stability gate. Across models
and datasets, stricter consensus decreases both coverage and exposed risk.
This makes the deployment choice reproducible. Operators select among measured
points generated by one shared evidence protocol.

\begin{table}[t]
  \centering
  \small
  \setlength{\tabcolsep}{3pt}
  \begin{tabular}{llrrrr}
    \toprule
    & & \multicolumn{2}{c}{IIIT5K} & \multicolumn{2}{c}{ICDAR13}\\
    \cmidrule(lr){3-4}\cmidrule(lr){5-6}
    Model & Setting & Cov. & MD@2 & Cov. & MD@2\\
    \midrule
    \multirow{3}{*}{LLaVA-Phi3}
      & Baseline & 100.0 & 33.7 & 100.0 & 35.6\\
      & \(m=1\) & 91.3 & 0.3 & 92.1 & 3.7\\
      & \(m=5\) & 67.7 & 0.0 & 68.3 & 0.9\\
    \multirow{3}{*}{Gemma3}
      & Baseline & 100.0 & 3.7 & 100.0 & 16.4\\
      & \(m=1\) & 95.8 & 0.0 & 95.1 & 1.8\\
      & \(m=5\) & 82.5 & 0.0 & 82.0 & 1.8\\
    \multirow{3}{*}{GLM-OCR}
      & Baseline & 100.0 & 3.7 & 100.0 & 1.8\\
      & \(m=1\) & 95.9 & 1.0 & 96.5 & 0.9\\
      & \(m=5\) & 90.5 & 0.0 & 91.0 & 0.9\\
    \bottomrule
  \end{tabular}
  \caption{Endpoint operating points under the shared \(K=5\) protocol.
  Coverage is percent and MD@2 is per mille.}
  \label{tab:endpoints}
\end{table}

Table~\ref{tab:endpoints} makes two properties visible. First, one shared
strictness order produces the expected coverage reduction on every backbone
and dataset. Second, each application can select a measured point according to
its coverage requirement. The \(m=5\) point nearly eliminates measured
catastrophic exposure and carries a larger coverage cost, especially for
LLaVA-Phi3. This ordered behavior confirms that \(m\) controls release
strictness consistently under the shared protocol.

\begin{figure}[t]
  \centering
  \includegraphics[width=\columnwidth]{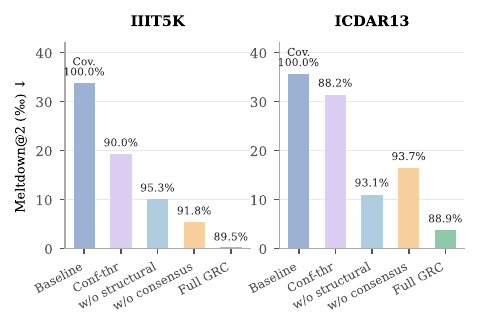}
  \caption{Component analysis on LLaVA-Phi3 at \(K=5,m=3\). Labels give
  coverage and bars give MD@2 (per mille). Consensus-only and structural-only
  controls both help. Their combination yields the lowest catastrophic
  exposure.}
  \label{fig:components}
\end{figure}

The component analysis in Fig.~\ref{fig:components} separates the two evidence
channels. Removing the structural screen retains cross-view vote and
dispersion control. Removing consensus retains structural screening. Each
channel suppresses catastrophic exposure, and the full controller achieves the
lowest rate. Approximately matched internal confidence remains substantially
weaker on the severe-error tail.

The ablation identifies complementary causal roles. Structural screening
detects outputs whose length or form conflicts with crop geometry. Consensus
detects formally valid outputs that change under controlled views. The
consensus-only variant serves as a direct multi-view comparator under the same
geometric protocol. Its gap from full GRC measures the added value of
structural evidence, and the structural-only gap measures the added value of
cross-view support.

\subsection{Query Cost and Deployment Role}

\begin{table}[t]
  \centering
  \small
  \setlength{\tabcolsep}{4pt}
  \begin{tabular}{lrrrrr}
    \toprule
    & & \multicolumn{2}{c}{IIIT5K} & \multicolumn{2}{c}{ICDAR13}\\
    \cmidrule(lr){3-4}\cmidrule(lr){5-6}
    \(K\) & Time & Cov. & MD@2 & Cov. & MD@2\\
    \midrule
    Baseline & 1.0x & 100.0 & 33.7 & 100.0 & 35.6\\
    3 & 2.7x & 85.4 & 0.0 & 86.3 & 1.8\\
    5 & 4.5x & 89.5 & 0.3 & 88.9 & 3.7\\
    7 & 6.3x & 89.5 & 0.3 & 89.1 & 3.7\\
    \bottomrule
  \end{tabular}
  \caption{LLaVA-Phi3 query-budget check at \(m=3\). Time is relative to one
  pass. Coverage is percent and MD@2 is per mille.}
  \label{tab:budget}
\end{table}

The query budget defines a measurable deployment envelope. A setting with
\(K=5\) costs 4.5x a single pass (Table~\ref{tab:budget}), and \(K=7\) yields
nearly identical coverage and MD@2 in this check. The selected default captures
the observed benefit before this saturation point. Independent views also
admit batching or parallel execution. A conditional routing design sends
ordinary crops through the primary OCR path and invokes GRC for uncertain or
high-value crops. Abstentions then enter review, a stronger recognizer, or a
task-specific verifier.

\section{Deployment Protocol}

GRC serves applications where unsupported text carries high downstream cost,
the backbone remains frozen, and abstentions have an escalation path. A primary
OCR path handles routine traffic, while policy routes uncertain or high-value
crops to GRC. Released candidates enter the downstream record with an
operating-point identifier. Abstained crops enter human verification, a
stronger OCR service, or a domain-specific checker. This routing structure
concentrates the multi-view cost on decisions where exposure control has clear
operational value.

Deployment proceeds from a labeled development set to a registered operating
policy. The operator first selects \(K\) from the latency envelope and \(m\)
from the measured risk--coverage curve. The complete protocol is then recorded
with the backbone and decoder versions. The transform family, normalization,
\(K_{\min}\), and \(\kappa\) then remain fixed for test and deployment. This
creates a named operating policy instead of ad hoc per-input thresholds.

Each invocation can retain its protocol identifier, transformed views,
canonical strings, validity flags, \(q\), \(\Delta\), and final decision.
These records support replay and inspection of the release rule, while
aggregate coverage, valid-view counts, consensus, and dispersion provide
compact monitoring signals. Material changes in these signals trigger protocol
re-evaluation and versioning instead of silent threshold adjustment, keeping
release decisions comparable across operating environments.

\section{Limitations and Future Work}

The present experiments focus on word and short-crop scene text under a fixed
family of geometric views. Broader languages, longer sequences, and
deployment-specific transform and latency calibration are natural extensions.
Future work can also combine GRC with complementary visual verification and
extend the protocol across a wider range of imaging conditions. Longer text can
use region-aware alignment before applying the same selective exposure
interface, while document workflows can route decisions at the field or region
level.

\section{Conclusion}

Reliable frozen-VLM OCR in audit-sensitive settings requires a decision about
whether generated text may be exposed. GRC supplies that decision through a
fixed black-box protocol. Controlled views provide external stability
evidence, a permissive structural screen removes degenerate continuations, and
predeclared operating points trade coverage and query cost for lower exposed
tail risk. Across all evaluated settings, this interface consistently reduces
mean, upper-tail, and catastrophic covered-output error. The broader lesson is
that when a plausible unsupported answer is worse than abstention, deployment
needs an exposure controller in addition to a capable generator.

The evidence supports this lesson from four complementary views. Main results
show the same risk-reduction direction across all six model--dataset settings.
The approximately coverage-matched comparison separates the value of external
geometric evidence from the amount of released output. Component analysis
identifies distinct structural and consensus contributions, while the query
sweep locates a practical operating budget. Together, these results connect the
decision rule to both its statistical effect and its deployment cost.

GRC is most valuable for frozen backbones when unsupported text is costly and
abstentions can be escalated. In this regime, one reproducible protocol makes
benefit, coverage, and query cost visible while providing a foundation for
longer-text control and complementary visual verification.

\bibliographystyle{plainnat}
\bibliography{grc}
\end{document}